\documentclass[sigconf]{acmart}
\AtBeginDocument{%
  }

\usepackage{CJKutf8}

\setcopyright{cc}
\copyrightyear{2026}
\acmYear{2026}
\acmDOI{10.1145/3776591.3836650}
\acmConference[ICMI Companion '26]{Companion of the INTERNATIONAL
  CONFERENCE ON MULTIMODAL INTERACTION}{October 05--09, 2026}{Napoli,
  Italy}
\acmBooktitle{Companion of the INTERNATIONAL CONFERENCE ON MULTIMODAL
  INTERACTION (ICMI Companion '26), October 05--09, 2026, Napoli, Italy}
\acmISBN{979-8-4007-2319-3/2026/10}

\begin{document}

\title{TEIDAN: A Multilingual Multiparty Dialogue Corpus}

% \author{Anonymous Author(s)}
% \affiliation{%
%   \institution{Anonymous Institution}
%   \country{}
% }
\author{Taiga Mori}
\affiliation{%
  \institution{Graduate School of Informatics, Kyoto University}
  \country{Japan}
}

\author{Koji Inoue}
\affiliation{%
  \institution{Graduate School of Informatics, Kyoto University}
  \country{Japan}
}

\author{Mikey Elmers}
\affiliation{%
  \institution{Graduate School of Informatics, Kyoto University}
  \country{Japan}
}

\author{Divesh Lala}
\affiliation{%
  \institution{Graduate School of Informatics, Kyoto University}
  \country{Japan}
}

\author{Tatsuya Kawahara}
\affiliation{%
  \institution{Graduate School of Informatics, Kyoto University}
  \country{Japan}
}

\renewcommand{\shortauthors}{Mori et al.}

\begin{abstract}
Multi-party interaction is a central setting for human communication and a
necessary target for human-agent interaction systems that must participate in
group conversation. Yet available corpora often focus on meetings,
task-oriented interaction, text-based interaction, or acted scenarios, and
fewer resources support cross-linguistic comparison of spontaneous face-to-face
triadic discussion. This paper presents TEIDAN, a multilingual multimodal
corpus that currently consists of Japanese and English three-party
conversations. TEIDAN records groups of three participants discussing
open-ended topics with individual pin microphones, a microphone array, and
participant-facing cameras, and provides IPU-based transcripts for both
language portions. Earlier studies used subsets of the Japanese portion for
task-specific benchmarks in multi-party dialogue modeling; in contrast, this
paper presents TEIDAN as a corpus resource spanning both Japanese and English,
with planned expansion to additional languages. We describe the collection
design, participants, recording setup, transcription format, and corpus
statistics, and provide preliminary analyses to illustrate how TEIDAN can
support research on turn-taking, addressee recognition, and multimodal
grounding in human-human and human-agent interaction.
\end{abstract}




\maketitle

\section{Introduction}
Dialogue systems have made rapid progress in dyadic interaction, but everyday
conversation rarely takes place in a strictly one-to-one channel. People
coordinate attention, stance, and floor rights in groups; they respond to
previous speakers, invite particular listeners to speak, and sometimes address
the group as a whole. These phenomena are especially important for spoken
dialogue systems and embodied agents that are expected to join human groups
without interrupting or ignoring the intended recipient of an utterance.

Prior work on multi-party interaction has shown that turn-taking depends on
both linguistic and non-linguistic cues, including lexical forms, gaze, body
orientation, and the local sequential organization of conversation. Existing
resources include text-based or online multiparty conversation
\cite{Reverdy2022RoomReader}, as well as corpora collected in meetings and
smart rooms \cite{Carletta2007AMI,Mostefa2007CHIL}, situated or task-oriented
interaction \cite{Kontogiorgos2018MutualGaze,Nihei2014Influential}, team or
game dialogue \cite{Litman2016Teams,Hung2010Wolf}, and human-robot interaction
\cite{Stefanov2016Focus}. These resources are valuable, but natural
face-to-face multiparty corpora with multimodal recordings remain limited,
especially when the interaction is open-ended rather than task-driven or
scenario-based. For Japanese, naturally occurring three-party casual
conversation has been studied using a chat corpus
\cite{Enomoto2020TurnTaking}. However, to our knowledge, there is still no
multilingual corpus that supports comparable analysis of spontaneous
face-to-face triadic interaction across languages.

The TEIDAN corpus was designed to address this gap.
Preliminary studies have used subsets of the Japanese portion as benchmarks for
addressee recognition, next-speaker prediction, and voice activity
projection~\cite{Inoue2025AddresseeCameraReady,Mori2026NextSpeakerCameraReady,elmers2025triadicCameraReady}.
These studies demonstrated that state-of-the-art language, multimodal, and
spoken-dialogue models still struggle with the social organization of
three-party talk.
However, they treated TEIDAN primarily as task data for model evaluation, and they did not provide a standalone description of the corpus as a multilingual multimodal resource.

This paper fills that role by presenting TEIDAN as a corpus in its own right.
For the Japanese portion, we follow the more recent corpus description and
statistics in prior work \cite{Mori2026NextSpeakerCameraReady}: the corpus contains 12
groups and 36 sessions. We then add a newly collected English portion designed
with a parallel triadic discussion protocol. The resulting corpus currently
contains 69 sessions, 57 unique participants, and 9 hours 46 minutes 57 seconds
of multimodal interaction. Our goal is not only to provide more data, but to
offer a comparable resource for examining which properties of multi-party
interaction are shared across languages and which may depend on language,
culture, or participant relationships. Although the current release focuses on
Japanese and English, TEIDAN is designed as an extensible multilingual corpus,
and future collections may add further languages under the same basic protocol.
The main contributions of this paper are:
\begin{itemize}
  \item the first corpus-level description of TEIDAN as a multilingual multimodal resource for spontaneous three-party discussion;
  \item a newly documented English extension that enables cross-linguistic and cross-cultural comparison with the Japanese portion;
  \item a summary of the multimodal recording setup, topic design,
  participant characteristics, transcript format, and corpus statistics; and
  \item preliminary quantitative and qualitative analyses illustrating how the
  corpus can support research on turn-taking, addressee recognition,
  next-speaker prediction, and multimodal grounding.
\end{itemize}

Table~\ref{tab:overview} provides a side-by-side overview of the two portions, highlighting the shared design and the language-specific properties discussed in the following sections.

\begin{table}[t]
  \caption{TEIDAN corpus overview. The upper section shows design choices shared
  across both portions; the lower section shows language-specific properties.}
  \label{tab:overview}
  \begin{tabular}{lcc}
    \toprule
    & Japanese & English \\
    \midrule
    Domain & \multicolumn{2}{c}{open-ended discussion} \\
    Modalities & \multicolumn{2}{c}{pin mic, array mic, video} \\
    Transcription unit & \multicolumn{2}{c}{IPU ($\geq$200\,ms)} \\
    Language & Japanese & English \\
    Participants (unique) & 24 & 33 \\
    Participant slots & 36 & 33 \\
    Participant relationship & acquainted & strangers \\
    Groups & 12 & 11 \\
    Topics per group & \multicolumn{2}{c}{3} \\
    Sessions & 36 & 33 \\
    Duration & 3:39:10 & 6:07:46 \\
    \bottomrule
  \end{tabular}
\end{table}

\section{Method}

\subsection{Corpus Design}

TEIDAN consists of face-to-face discussions among triads.
Each group contains three participants seated around a round table.
The design intentionally avoids a strict task objective: participants are given broad prompts and are asked to discuss their opinions rather than solve a game or produce a single correct answer.
This design follows the corpus description in prior work~\cite{Mori2026NextSpeakerCameraReady}: unlike meeting corpora or task-oriented interaction data, TEIDAN captures spontaneous opinion exchange where floor management and participation emerge from the interaction itself.

The Japanese corpus covers six open-ended topics~\cite{Mori2026NextSpeakerCameraReady}: ``If Japan were to relocate its capital, where would it be?''~(city); ``If you could bring only one item to a deserted island, what would it be?''~(island); ``Where would you go if you were to travel this week?''~(travel); ``For a day off, would you go to the sea, mountains, or city?''~(outdoor); ``What is the most important thing in life?''~(life); and ``How would you travel between Kyoto and Tokyo?''~(trans).
The English corpus uses three open-ended topics corresponding to topics in the Japanese corpus: ``If you could bring only one item to a deserted island, what would it be?''~(island); ``Where would you go if you were to travel, and what would you do there?''~(travel); and ``What is more important to you: family, life experiences, or money?''~(life).
These three shared topic themes support cross-linguistic comparison under comparable discussion prompts.

\subsection{Participants}

The Japanese corpus was collected from students and faculty members in the same
laboratory.
The 12 groups were formed from 24 unique participants, so some individuals appear in more than one group.
The participants were therefore already acquainted with one another before recording, and many pairs share prior interaction history before the sessions.
This familiarity is an important property of the Japanese data: the conversations include interaction among people who share an institutional context and can rely on prior interpersonal knowledge.

The English corpus was collected from foreign-national residents of Japan who
were recruited through a staffing agency. Before recording, they completed a
questionnaire covering age, gender, country of origin, and language background.
The participants were native or near-native speakers of English and met one
another for the first time at the recording. They represented 13 nationalities:
USA (13), India (3), Philippines (3), Malaysia (3), UK (2), South Africa (2),
Argentina (1), Australia (1), Canada (1), Ireland (1), New Zealand (1),
Singapore (1), and Turkey (1). Thus, the English data differs from the Japanese
data not only in language and cultural background, but also in the
participants' pre-existing relationship. This contrast is part of the corpus
design, but it also means that cross-linguistic comparisons should be
interpreted carefully: observed differences may reflect participant
relationship and recruitment context as well as language or cultural
background. Nevertheless, the questionnaire metadata make it possible to
examine interactional patterns in relation to speakers' diverse backgrounds
and to conduct detailed analyses within the English portion.

\subsection{Recording Setup}
\begin{figure}[t]
  \centering
  \includegraphics[width=\columnwidth]{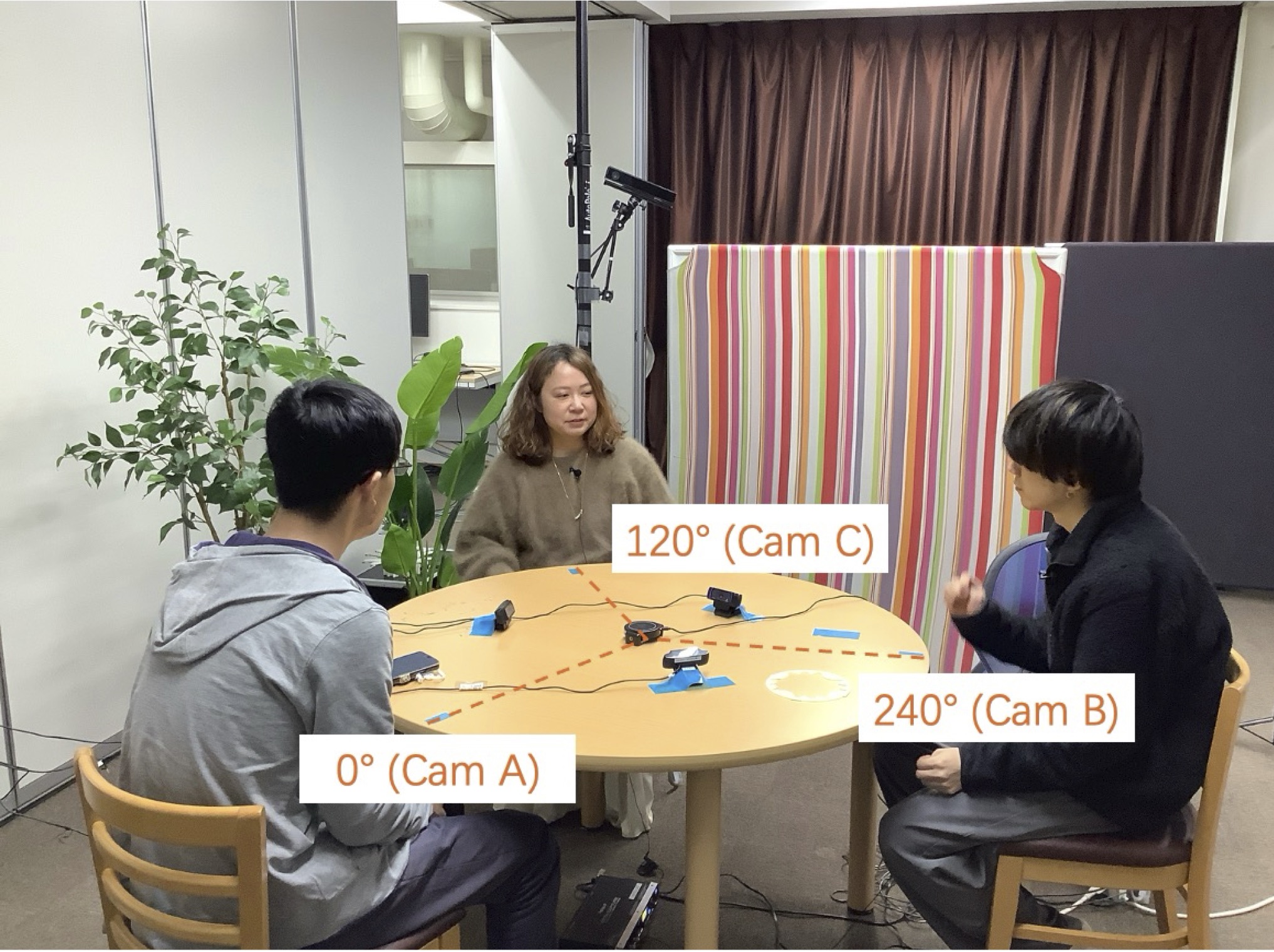}\\
  \small (a) Top view\\[0.6em]
  \includegraphics[width=\columnwidth]{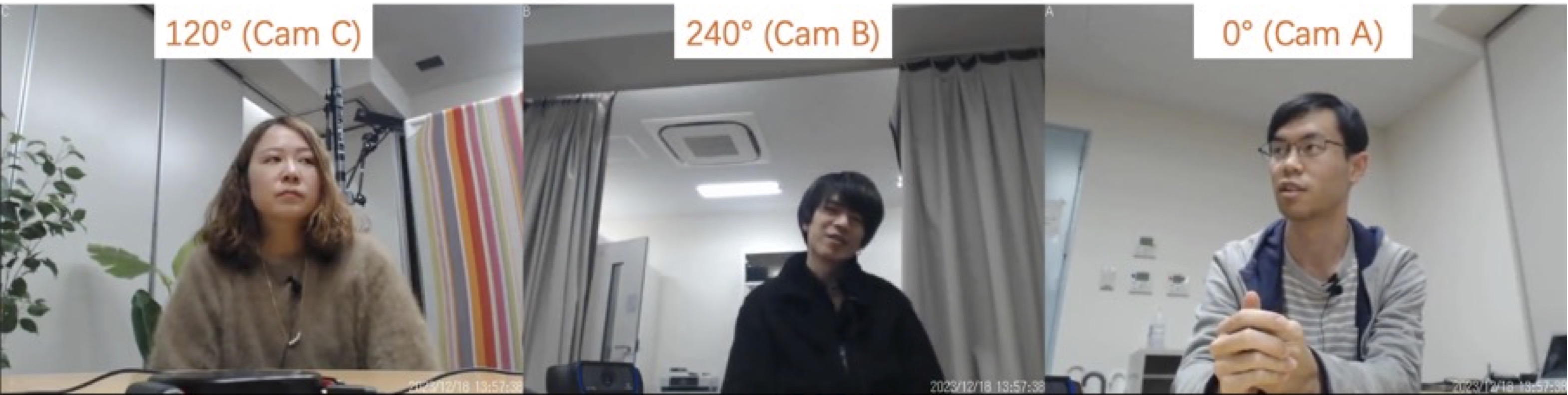}\\
  \small (b) Front view
  \caption{Recording setup used for TEIDAN.}
  \Description{Top and front views of the TEIDAN recording setup, showing three participants seated around a round table with microphones and cameras.}
  \label{fig:recording-setup}
\end{figure}

Both language portions use a multimodal recording setup. Participants wear
individual pin microphones so that each speaker's speech can be captured
separately. A microphone array records the shared acoustic scene, and cameras
record the participants' faces. Figure~\ref{fig:recording-setup} shows the
recording layout from top and front views. In the Japanese setup, each participant is
recorded by a participant-facing camera and individual pin microphone
\cite{Inoue2025AddresseeCameraReady,Mori2026NextSpeakerCameraReady}. Documentation for the corpus also
records the direction-of-arrival (DOA) configuration of the microphone array:
for sessions 01 to 12, DOA labels at 0, 120, and 240 degrees correspond to
cameras A, C, and B, respectively. These labels are intended to support source
separation and multimodal alignment.

The English corpus was collected on September 18 and 19 and September 25, 2025.
Before recording, participants completed a language background questionnaire
and signed a consent form granting permission to publicly release, transfer,
or license the recorded audio and video data for research on speech
recognition, enhancement, and dialogue systems. The consent form included
opt-out options allowing individual participants to restrict redistribution of
video, or of both audio and video. Eleven groups were recorded, with 33 unique
participants in total. Each group discussed three topics in a randomized order,
yielding 33 sessions.

\subsection{Transcription}
The corpus follows a transcription format based on the Corpus of Spontaneous
Japanese (CSJ) \cite{maekawa2003corpus}, and the same convention is applied to
both language portions. An inter-pausal unit (IPU) is a speech segment bounded
by pauses of at least 200\,ms. When a pause of 200\,ms or more occurs within a
word, the unit boundary is not split; instead, the pause position is recorded
with a P~tag.
Non-linguistic vocalizations, namely laughter (\texttt{\{LAUGH\}}) and coughing
(\texttt{\{COUGH\}}), constitute independent IPUs even when temporally adjacent
to speech. Each IPU entry records start and end times in seconds and a speaker
label (A, B, or C).

Japanese transcripts provide two parallel columns per line: a base orthographic
form and a phonetic form in katakana, separated by \texttt{\&}. English
transcripts record the base form only. Both languages share the same set of
inline tags: backchannels \texttt{(I~\ldots)}, fillers \texttt{(F~\ldots)},
self-interruptions \texttt{(D~\ldots)}, and uncertain or inaudible portions
\texttt{(?~\ldots)}. Participant-identifying proper nouns are anonymized with an
N~tag. Both the Japanese and English corpora have been fully transcribed using
this format.

Session identifiers encode language, group, and topic. For
example, \texttt{J01\_city} denotes the Japanese conversation from group 01 on
the city topic. English sessions use the same convention with the language
prefix \texttt{E}. Within each session, participants are labeled A, B, and C.

\section{Statistics}
The Japanese statistics follow the 12-group subset used in prior work
\cite{Mori2026NextSpeakerCameraReady}. The 12 groups were drawn from 24 unique
participants, with some individuals appearing in more than one group. Each
group participated in three approximately 5 to 7 minute sessions, each
corresponding to one topic, yielding 36 sessions and 3 hours 39 minutes 10
seconds of data. The English portion
contains 11 groups and 33 unique participants. Each English group discussed
three topics, yielding 33 sessions. Table~\ref{tab:utterance-stats} lists the
per-session durations and IPU counts for both language portions.

\begin{table*}[t]
  \caption{Per-session duration and IPU counts. The three values in the IPU
  column give speaker A, B, and C counts respectively.}
  \label{tab:utterance-stats}
  \begin{tabular}{lcc|lcc}
    \toprule
    Session & Time & IPU (A/B/C) & Session & Time & IPU (A/B/C) \\
    \midrule
    \multicolumn{6}{l}{\textit{Japanese (36 sessions)}} \\
    J01\_city   & 6:14 & 65/81/137   & J07\_life    & 5:34 & 97/79/150   \\
    J01\_island & 6:37 & 89/112/134  & J07\_outdoor & 5:43 & 114/97/155  \\
    J01\_travel & 8:10 & 97/99/165   & J07\_trans   & 5:53 & 98/99/185   \\
    J02\_city   & 5:51 & 76/94/98    & J08\_life    & 6:55 & 98/93/101   \\
    J02\_island & 6:22 & 119/133/94  & J08\_outdoor & 5:40 & 85/74/87    \\
    J02\_travel & 5:20 & 96/106/110  & J08\_trans   & 5:58 & 60/62/107   \\
    J03\_city   & 6:15 & 81/123/128  & J09\_life    & 6:40 & 104/113/147 \\
    J03\_island & 7:28 & 106/161/156 & J09\_outdoor & 5:30 & 88/118/139  \\
    J03\_travel & 5:41 & 115/129/118 & J09\_trans   & 5:41 & 86/117/148  \\
    J04\_city   & 5:48 & 146/142/123 & J10\_life    & 5:37 & 96/87/149   \\
    J04\_island & 5:51 & 138/126/153 & J10\_outdoor & 5:48 & 118/81/161  \\
    J04\_travel & 5:39 & 115/116/94  & J10\_trans   & 5:52 & 117/81/151  \\
    J05\_city   & 5:19 & 108/119/95  & J11\_life    & 6:27 & 96/124/135  \\
    J05\_island & 5:30 & 121/164/101 & J11\_outdoor & 6:39 & 139/148/122 \\
    J05\_travel & 5:01 & 100/144/110 & J11\_trans   & 5:47 & 133/122/140 \\
    J06\_city   & 6:13 & 86/92/93    & J12\_life    & 6:29 & 79/86/149   \\
    J06\_island & 6:17 & 127/113/137 & J12\_outdoor & 5:41 & 108/106/135 \\
    J06\_travel & 6:13 & 112/75/143  & J12\_trans   & 7:10 & 150/111/144 \\
    \midrule
    \multicolumn{6}{l}{\textit{English (33 sessions)}} \\
    E01\_life   & 11:50 & 247/138/113 & E07\_life   & 11:13 & 225/125/152 \\
    E01\_island & 10:26 & 255/132/99  & E07\_island & 9:01  & 229/161/172 \\
    E01\_travel & 10:26 & 194/127/159 & E07\_travel & 10:08 & 237/160/142 \\
    E02\_life   & 10:25 & 170/207/232 & E08\_life   & 11:39 & 108/119/237 \\
    E02\_island & 10:43 & 208/197/240 & E08\_island & 12:15 & 98/142/260  \\
    E02\_travel & 10:17 & 165/222/223 & E08\_travel & 11:59 & 127/173/248 \\
    E03\_life   & 11:05 & 142/92/136  & E09\_life   & 12:23 & 130/163/209 \\
    E03\_island & 9:55  & 151/96/196  & E09\_island & 10:15 & 151/149/178 \\
    E03\_travel & 12:02 & 178/90/212  & E09\_travel & 11:17 & 192/153/183 \\
    E04\_life   & 11:12 & 151/75/155  & E10\_life   & 11:13 & 170/172/129 \\
    E04\_island & 11:31 & 160/230/218 & E10\_island & 10:30 & 178/204/165 \\
    E04\_travel & 10:01 & 128/125/177 & E10\_travel & 11:56 & 186/192/160 \\
    E05\_life   & 12:11 & 188/235/278 & E11\_life   & 10:33 & 173/124/179 \\
    E05\_island & 10:58 & 191/269/345 & E11\_island & 11:51 & 148/164/221 \\
    E05\_travel & 11:20 & 195/278/280 & E11\_travel & 12:44 & 142/139/163 \\
    E06\_life   & 10:21 & 89/89/142   &             &       &             \\
    E06\_island & 10:31 & 203/166/210 &             &       &             \\
    E06\_travel & 13:19 & 200/123/217 &             &       &             \\
    \midrule
    Avg.\ (J)   & 6:05  & 344         & Total (J)   & 3:39:10 & 12,384   \\
    Avg.\ (E)   & 11:08 & 526         & Total (E)   & 6:07:46 & 17,370   \\
    Avg.\ (all) & 8:30  & 431         & Total (all) & 9:46:57 & 29,754   \\
    \bottomrule
  \end{tabular}
  \vspace{1em}
  \caption{Corpus-level IPU and pseudo-turn statistics after excluding
  backchannel and non-linguistic IPUs. All duration values are in seconds.}
  \label{tab:prelim-turn}
  \begin{tabular}{lrrrrrrrrr}
    \toprule
    & \multicolumn{4}{c}{IPU} & \multicolumn{5}{c}{Pseudo-turn} \\
    \cmidrule(lr){2-5} \cmidrule(lr){6-10}
    Language & $n$ & Mean & Min & Max & $n$ & Mean & Min & Max & IPUs/turn \\
    \midrule
    English  & 10,097 & 1.93 & 0.063 & 12.2 & 4,851 & 4.65 & 0.065 & 188.0 & 2.08 \\
    Japanese & 7,348  & 1.67 & 0.127 & 18.6 & 4,748 & 2.95 & 0.140 & 97.4  & 1.55 \\
    \bottomrule
  \end{tabular}
\end{table*}

\section{Preliminary Analysis}

\subsection{Quantitative Analysis}
To illustrate the kinds of analyses supported by TEIDAN, we conducted a
preliminary comparison of the organization of turns and turn-taking in the
Japanese and English portions.
This analysis provides a descriptive comparison of the two corpus portions; it
does not treat individual IPUs or pseudo-turns as independent samples for
inferential testing.
As described above, the TEIDAN transcripts mark backchannels with the
\texttt{I} tag and non-linguistic vocalizations such as laughter and coughing
with dedicated tags. Because our focus is on floor-holding turns, we first
excluded IPUs consisting of backchannels or non-linguistic vocalizations. For
each session, we then sorted the remaining IPUs by start time and merged each
sequence of consecutive IPUs produced by the same speaker into a single
pseudo-turn. Thus, after filtering, an intervening IPU affects the pseudo-turn
boundary only when it is produced by a different speaker. This procedure
reduced the English portion from 17,370 total IPUs to 10,097 retained IPUs and
then to 4,851 pseudo-turns; the corresponding counts for the Japanese portion
were 12,384 total IPUs, 7,348 retained IPUs, and 4,748 pseudo-turns.
Table~\ref{tab:prelim-turn} summarizes the resulting IPU and pseudo-turn
statistics.

The English portion contains more than 1.3 times as many substantive IPUs
as the Japanese portion (10,097 vs. 7,348), but the number of pseudo-turns is
almost the same across languages (4,851 vs. 4,748). This pattern is likely
related to the number of IPUs that constitute each turn: English pseudo-turns
contain more than 1.3 times as many IPUs on average as Japanese pseudo-turns
(2.08 vs. 1.55). As a result, English turns are also more than 1.5 times longer
on average than Japanese turns (4.65 vs. 2.95 seconds). The maximum English
pseudo-turn lasts 188 seconds, indicating that the corpus includes cases where
a single speaker holds the floor for more than three minutes. These preliminary
results show a clear difference in turn organization between the English and
Japanese portions of TEIDAN. Because IPUs and pseudo-turns produced within the
same group are interactionally related, the pooled statistics should be
interpreted as corpus-level descriptive differences rather than as an
inferential test of population-level differences between English and Japanese
conversation.

\subsection{Qualitative Analysis}

We further provide qualitative examples to show how the corpus can support
close analysis of turn construction in multimodal multiparty interaction. We
first show a typical case observed in the English portion. In the following
excerpt, the participants are discussing what is most important in life.
However, as noted above, backchannels are excluded.\\

{\setlength{\parindent}{0pt}
Excerpt 1. E06\_life 6:26.414--9:37.017\\
\hspace*{1.0em}01 A: … if you're lucky, life experiences,\\
\hspace*{1.0em}02 \hspace*{1.1em}I think life experiences and/or family,\\
\hspace*{1.0em}03 \hspace*{1.1em}I think are, are your, your main goals.\\
\hspace*{1.0em}04 \hspace*{1.1em}Or they are, if you're lucky anyway.\\
\hspace*{1.0em}05 \hspace*{1.1em}If you, if you don't have to worry that much about \\
\hspace*{1.0em}06 \hspace*{1.1em}putting a roof over your head or your kid's head, \\
\hspace*{1.0em}07 \hspace*{1.1em}(L whatever L) (?). \\
\hspace*{1.0em}08 C: It’s interesting, you know, \\
\hspace*{1.0em}09 \hspace*{1.1em}you say, some people say, like, \\
\hspace*{1.0em}10 \hspace*{1.1em}you’re the sum of all your memories. … \\
\hspace*{1.0em}11 \hspace*{1.1em}And that's where the idea of family comes in. … \\
\hspace*{1.0em}12 \hspace*{1.1em}I wanna share my entire life with these people. \\
\hspace*{1.0em}13 \hspace*{1.1em}And that itself gives you more joys. … \\
\hspace*{1.0em}14 \hspace*{1.1em}It's more the everyday life that you're building \\
\hspace*{1.0em}15 \hspace*{1.1em}towards the relationship or the stronger bond \\
\hspace*{1.0em}16 \hspace*{1.1em}with these individuals \\
\hspace*{1.0em}17 \hspace*{1.1em}who are gonna outlive you and all your memories.\\
}

In the larger turn from which lines 01 to 07 are excerpted, participant A
describes, based on their own life, why they place the greatest value on life
experiences. While A is developing this opinion, the other participants do not
take the floor, although brief backchannels may occur; A therefore holds the
floor from the beginning to the end of the opinion statement. This is the
longest pseudo-turn reported above, lasting 187.956 seconds. The next speaker
is C. At the beginning of C's turn, C comments on A's preceding opinion with
``It's interesting,'' and then continues by presenting their own view. The other
participants likewise do not take the floor during C's opinion statement, so C
also holds the floor for 172.594 seconds.
In this way, the English conversations often show a pattern in which one
participant takes a relatively long turn to present an opinion, while the other
participants rarely take the floor during that turn, regardless of whether they
agree or disagree. After one participant has presented an opinion, the next
speaker may begin their own turn by briefly commenting on the preceding opinion
before moving into their own contribution.

The next excerpt illustrates a typical pattern in the Japanese portion. Here,
too, the participants are discussing what is most important in life.\\

{\setlength{\parindent}{0pt}
\begin{CJK}{UTF8}{min}
Excerpt 2. J09\_life 1:10.550--1:39.220\\
\hspace*{1.0em}01 C: なるほど(N 吉田)さんはどうですか\\
\hspace*{3.2em}\emph{I see. How about you, Yoshida-san?}\\
\hspace*{1.0em}02 A: (F まー)難しいですけど(F まー)お金\\
\hspace*{3.2em}\emph{Well, it is difficult, but, well, money}\\
→03 C: やっぱり\\
\hspace*{3.2em}\emph{After all.}\\
\hspace*{1.0em}04 A: がやっぱり大事かなとは思いますけど\\
\hspace*{3.2em}\emph{is probably important after all, I think.}\\
→05 C: (F ま)何だかんだ言ってね確かに\\
\hspace*{3.2em}\emph{Well, when all is said and done, yes, certainly.}\\
\hspace*{1.0em}06 A: (F まー)お金がないと(F まー)\\
\hspace*{3.2em}\emph{Well, if you do not have money, well,}\\
\hspace*{1.0em}07 \hspace*{1.1em}留学とかもできない\\
\hspace*{3.2em}\emph{you cannot even do things like study abroad}\\
→08 B: (L (I まあ) L)(L それ言われたらそうですね L)\\
\hspace*{3.2em}\emph{Well, if you put it that way, that is true.}\\
→09 C: 確かにね確かに\\
\hspace*{3.2em}\emph{That's true, yes, true.}\\
\hspace*{1.0em}10 A: なってしまうんで\\
\hspace*{3.2em}\emph{it ends up being like that, so}\\
\hspace*{1.0em}11 \hspace*{1.1em}(F まー)(F そのー)なんというか\\
\hspace*{3.2em}\emph{well, um, how should I put it,}\\
\hspace*{1.0em}12 C: そっか\\
\hspace*{3.2em}\emph{I see.}\\
\hspace*{1.0em}13 A: めちゃくちゃ(F その)富を集める必要はない\\
\hspace*{3.2em}\emph{I do not think it is necessary to accumulate}\\
\hspace*{1.0em}14 \hspace*{1.1em}とは思うんですけど\\
\hspace*{3.2em}\emph{an extreme amount of wealth, but}\\
\hspace*{1.0em}15 \hspace*{1.1em}やっぱり(F その)生きていくには\\
\hspace*{3.2em}\emph{after all, in order to live,}\\
\hspace*{1.0em}16 \hspace*{1.1em}(D ヒツヨ)最低限は必要というか\\
\hspace*{3.2em}\emph{you need, or rather, at least the minimum.}\\
\end{CJK}
}

Immediately before this excerpt, B had stated that life experiences were the
most important thing, using their own short-term study abroad experience as an
example. In response, C asks A for an opinion in line 01. A begins to state
their opinion in line 02. At the end of line 02, A says `\begin{CJK}{UTF8}{min}お金\end{CJK}' (`money')
while shifting their gaze from no specific recipient to C. Line 02 could be
grammatically complete at that point. However, given that A is the youngest
participant in the group and uses polite forms elsewhere, ending the utterance
there would sound too abrupt; one might expect a polite ending such as
`\begin{CJK}{UTF8}{min}だと思います\end{CJK}' (`I think'). Thus, A's utterance still sounds incomplete. The final syllable of
`\begin{CJK}{UTF8}{min}お金\end{CJK}' (`money') is also lengthened. Given that the utterance is still in progress,
this can be interpreted as a try-marking-like practice
\cite{sacks1979two,schegloff2007sequence}. What appears to be tested here,
however, is not C's recognition of a referent, but rather C's stance toward the
opinion that money is important.

Indeed, in line 03, C takes the turn with `\begin{CJK}{UTF8}{min}やっぱり\end{CJK}' (`after all'). This
expression displays an anticipation of A's view, and C behaves cooperatively
toward A, although not necessarily by offering full agreement. This also
suggests that C treats A as seeking C's stance. A then resumes the turn in line
04 in a way that syntactically continues from line 02, completing the utterance.
In response, C displays partial agreement with A in line 05. During this
sequence, B does not yet show any clearly cooperative response to A. A then
takes the turn again from line 06. At this point, A shifts their gaze from no
specific recipient to B and says that without money one cannot even study
abroad, thereby referring to B's immediately preceding experience while
emphasizing the importance of money. A also appears to produce beat gestures,
soliciting agreement from B. In fact, immediately after A finishes line 07, B
partially agrees with A for the first time in line 08. C then also agrees with
A again in line 09. As in line 02, line 07 is grammatically complete, but it
again sounds too abrupt as a possible turn ending; for example, a polite expression such as
`\begin{CJK}{UTF8}{min}じゃないですか?\end{CJK}' (`isn’t that so?') could be expected after it. Thus, A is still projected to
continue speaking. After B and C agree with A, A resumes the turn from line 10
in a way that syntactically connects to line 07 and completes the statement of
their opinion.

In this way, the Japanese discussions often include cases in which participants
present their opinions step by step, moving forward while seeking agreement or
empathy from other participants along the way. Participants also use cultural
norms such as polite-form endings, together with syntactic structure, to project
that they will take the turn again while creating space for other participants
to respond.

The next excerpt is another typical example from the Japanese portion. In this
excerpt, the participants are discussing where Japan's capital should be moved
if it were relocated from Tokyo.\\

{\setlength{\parindent}{0pt}
\begin{CJK}{UTF8}{min}
Excerpt 3. J01\_city 2:51.380--3:49.560\\
\hspace*{1.0em}01 A: いや(F あの)僕大阪がいいって思ってる点が\\
\hspace*{3.2em}\emph{Well, um, the reason I think Osaka is good is}\\
\hspace*{1.0em}02 \hspace*{1.1em}もう一つあって\\
\hspace*{3.2em}\emph{there is one more point,}\\
\hspace*{1.0em}03 \hspace*{1.1em}それが(F あの)\\
\hspace*{3.2em}\emph{and that is, um,}\\
\hspace*{1.0em}04 \hspace*{1.1em}都市計画\\
\hspace*{3.2em}\emph{the city planning,}\\
\hspace*{1.0em}05 \hspace*{1.1em}道路道路や鉄道網ってのが\\
\hspace*{3.2em}\emph{the roads, roads and railway networks,}\\
\hspace*{1.0em}06 \hspace*{1.1em}非常に直線的で分かりやすい\\
\hspace*{3.2em}\emph{are very straight and easy to understand,}\\
\hspace*{1.0em}07 \hspace*{1.1em}構造をしているっていうのが\\
\hspace*{3.2em}\emph{that they have that kind of structure.}\\
→08 C: (F あ)御堂筋とか\\
\hspace*{3.2em}\emph{Ah, like Midosuji?}\\
\hspace*{1.0em}09 A: そうです\\
\hspace*{3.2em}\emph{Exactly.}\\
\hspace*{1.0em}10 \hspace*{1.1em}(F あのー)東京の地下鉄の路線図(F あの)\\
\hspace*{3.2em}\emph{Um, the Tokyo subway route map, um,}\\
\hspace*{1.0em}11 \hspace*{1.1em}ご覧になったことご覧になったこと\\
\hspace*{3.2em}\emph{you have seen it, you have seen it,}\\
\hspace*{1.0em}12 \hspace*{1.1em}あると思うんですけど\\
\hspace*{3.2em}\emph{I think you probably have, but}\\
\hspace*{1.0em}13 \hspace*{1.1em}(F えーっと)かなりもう\\
\hspace*{3.2em}\emph{Uh, it is already quite}\\
\hspace*{1.0em}14 \hspace*{1.1em}ぐねぐねと曲がってて\\
\hspace*{3.2em}\emph{winding and twisting,}\\
\hspace*{1.0em}15 \hspace*{1.1em}複雑怪奇じゃないですか\\
\hspace*{3.2em}\emph{and extremely complicated, right?}\\
\hspace*{1.0em}16 \hspace*{1.1em}そう(F う)そうじゃなく(F えっと)(D ス)(D ス)\\
\hspace*{3.2em}\emph{Right, uhm, not like that, uh, s-, s-,}\\
\hspace*{1.0em}17 \hspace*{1.1em}一方で(F あのー)\\
\hspace*{3.2em}\emph{whereas, um,}\\
\hspace*{1.0em}18 \hspace*{1.1em}大阪はこう縦横に揃った\\
\hspace*{3.2em}\emph{Osaka is, like, aligned vertically and horizontally,}\\
→19 C: (F あー)なんとか筋(D ナン)\\
\hspace*{3.2em}\emph{Ah, something-suji, some-}\\
→20 \hspace*{1.1em}なんとか筋線みたいな\\
\hspace*{3.2em}\emph{like some-suji line.}\\
\hspace*{1.0em}21 A: (F えーっと)そう\\
\hspace*{3.2em}\emph{Um, yes,}\\
\hspace*{1.0em}22 \hspace*{1.1em}そういった都市計画をしている…\\
\hspace*{3.2em}\emph{it has that kind of city planning…}\\
\hspace*{1.0em}23 \hspace*{1.1em}…やはりこう都市として\\
\hspace*{3.2em}\emph{…after all, as a city,}\\
\hspace*{1.0em}24 \hspace*{1.1em}優れているところかなと思います\\
\hspace*{3.2em}\emph{what makes it superior, I think.}\\
\end{CJK}
}

From line 01, A begins to present their opinion that Osaka would be a good
choice. By line 07, A has given as a reason that Osaka's roads and railways are
linear and easy to understand. At line 07, A's utterance is syntactically
incomplete and is clearly not a transition-relevance place (TRP)
\cite{sacks1974simplest}. Nevertheless, C takes the turn in line 08, saying
`\begin{CJK}{UTF8}{min}御堂筋とか\end{CJK}' (`like Midosuji?').
Midosuji refers to a straight north-south road and subway line in Osaka. C's
utterance is formally a confirmation question, and A indeed responds with
`\begin{CJK}{UTF8}{min}そうです\end{CJK}' (`Exactly') in line 09. While
responding to C, A resumes the turn. After referring to the Tokyo railway map,
A contrasts it with Osaka's railway lines. Immediately after A says
`\begin{CJK}{UTF8}{min}大阪はこう縦横に揃った\end{CJK}' (`Osaka is, like,
aligned vertically and horizontally') in line 18, C again takes the turn in line
19 at a position that is not a TRP. In lines 19 and 20, C says
`\begin{CJK}{UTF8}{min}なんとか筋線みたいな\end{CJK}' (`like some-suji line').
Osaka's urban layout is often described as a grid, with roads running straight
north-south and east-west; the north-south roads are called \emph{suji}. As
in line 08, C's utterance is formally a confirmation question. In line 21, A
responds with `\begin{CJK}{UTF8}{min}そう\end{CJK}' (`yes') and then
continues to present their opinion to the end.

Why, then, does C take the turn at these points, even though A's utterance has
not reached a TRP? The simplest interpretation is that, because A has not named
specific roads or railway lines, C asks a question in order to check their own
understanding. However, C is from Osaka and is therefore expected to know the
geography of Osaka better than the other participants. Moreover, the intonation
of C's utterances in lines 08 and 19-20 is falling, or at least does not clearly
rise. Given this, C's utterances can also be understood not as questions asked
for C's own understanding, but as concrete examples that support A's claim, or
as resources for helping B, the third participant, understand A's point. In
fact, the utterances in lines 08 and 19-20 are maximally compact, minimizing the
interruption to A's ongoing turn. In the video, immediately after these
utterances, B can also be seen nodding while looking at A, displaying
understanding.

Seen in this way, C's utterances in lines 08 and 19--20 are not intrusive
interruptions into A's turn, but rather cooperative actions oriented to A and
A's ongoing activity. Thus, in Japanese conversations, listeners sometimes take
brief turns even at non-TRP positions in order to support the current speaker.

\section{Discussion}
The preliminary analyses illustrate how TEIDAN can support cross-linguistic
and multimodal investigation of turn-taking. The quantitative comparison shows
a clear difference between the English and Japanese portions in the number of
IPUs per pseudo-turn and the mean duration of pseudo-turns. The magnitude of
the observed difference is substantial at the level of the pooled corpus
statistics, although the present descriptive analysis does not quantify
variation between groups or support population-level inference. The
qualitative examples then show how such differences can be examined in detail:
the English excerpt illustrates extended floor-holding during opinion
statements, whereas the Japanese excerpts show compact listener contributions
embedded in the ongoing development of another participant's turn. Thus, the
corpus enables researchers to move between corpus-level statistics and close
sequential analysis of multimodal interaction.

These observations suggest that language and culture may affect the
organization of turns and turn-taking. In particular, the Japanese examples
suggest a more collaborative organization of participation: speakers build
their opinions incrementally, and listeners provide compact responses,
candidate understandings, or supportive examples before the main speaker's turn
has reached completion. This does not simply mean that Japanese participants
speak less. Rather, the shorter pseudo-turns observed in the Japanese portion
may reflect an interactional orientation toward maintaining alignment with
co-participants.

This pattern is consistent with broader accounts of Japanese interaction as
high-context communication, in which shared background, timing, and implicit
coordination play an important role \cite{hall1976beyond}. It is also
compatible with work on cultural models of the self, which characterizes
Japanese culture as emphasizing interdependent self-construal and attention to
relations with others \cite{markus1991culture}. From this perspective, the
Japanese speakers and listeners appear to jointly shape the progress of an
opinion statement, with listeners contributing in ways that support the
speaker's ongoing activity while keeping their own turns compact.

At the same time, these observations should not be interpreted as isolating
language or culture as independent causal factors. The Japanese and English
portions differ not only in language, but also in participant relationship: the
Japanese participants were acquainted members of the same laboratory, whereas
the English participants were strangers recruited for the recording. These
differences may also affect how actively listeners enter an ongoing turn or how
long speakers hold the floor. TEIDAN's value is that it makes such comparisons
observable in a shared multimodal format, enabling future work to examine how
linguistic resources, embodied behavior, participant relationships, and
cultural norms jointly shape multi-party interaction.

\section{Conclusion}
We presented TEIDAN, a multilingual multimodal corpus of spontaneous
three-party dialogue in Japanese and English. Unlike prior studies that used
subsets of the Japanese data for task-specific benchmarks, this paper provides
a corpus-level account of the collection protocol, participant characteristics,
recording setup, transcription format, and statistics across both language
portions. TEIDAN currently contains 69 sessions of open-ended face-to-face
discussion with individual audio, array audio, video, and IPU-based
transcripts.

The preliminary analyses illustrated the kinds of questions that TEIDAN can
support, linking corpus-level statistics with close sequential analysis of
turn-taking and multimodal listener behavior. These analyses should be read as
demonstrations of the corpus's potential rather than as definitive claims about
language or culture, especially because the Japanese and English portions also
differ in participant relationship. Future work will enrich the corpus with
annotations for conversational phenomena such as addressees and
transition-relevance places (TRPs), making it applicable to a wider range of
model-building tasks for multi-party dialogue and human-agent interaction. We
are also considering releasing the dataset under conditions that respect
participant consent and privacy, and plan to expand the corpus to additional
languages.

\begin{acks}
This work was supported by JST Moonshot R\&D JPMJPS2011 and JST PRESTO JPMJPR24I4.

\end{acks}

\section*{Safe and Responsible Innovation Statement}
The corpus contains recordings of human participants and therefore requires
careful handling of privacy and consent. All participants signed consent forms
before recording, granting permission to publicly release, transfer, or license
the recorded audio and video data for research on speech recognition,
enhancement, and dialogue systems. The consent form included opt-out options
allowing individual participants to restrict redistribution of video, or of
both audio and video; corpus release must respect these per-participant
restrictions. Public examples should be anonymized where possible, and any use
beyond the stated research purpose should be separately disclosed. Models trained
or evaluated on TEIDAN should be analyzed for potential differences across
language groups, because cross-cultural comparison can otherwise reinforce
overgeneralized assumptions about conversational behavior.

\bibliographystyle{ACM-Reference-Format}
\bibliography{sample-base}

\end{document}